\documentclass{article}
\usepackage{NDI}

\usepackage{times}
\usepackage{soul}
\usepackage{url}
\usepackage[utf8]{inputenc}
\usepackage[small]{caption}
\usepackage{graphicx}
\usepackage{amsmath}
\usepackage{amsthm}
\usepackage{booktabs}
\usepackage{algorithm}
\usepackage{algorithmic}
\usepackage{amssymb}

\title{Absolute Wrong Makes Better: Boosting Weakly Supervised Object Detection via Negative Deterministic Information}

\author{
Guanchun Wang$^1$,
Xiangrong Zhang$^1$\footnote{Contact Author},
Zelin Peng$^{1}$,
Xu Tang$^1$,
Huiyu Zhou$^2$\And
Licheng Jiao$^1$
\affiliations
$^1$Xidian University\\
$^2$University of Leicester\\
\emails
\{gwang\_2, zlpeng\}@stu.xidian.edu.cn, \{xrzhang, lchjiao\}@mail.xidian.edu.cn,\\
tangxu128@gmail.com, hz143@leicester.ac.uk
}

\begin{document}

\maketitle

\begin{abstract}
Weakly supervised object detection (WSOD) is a challenging task, in which image-level labels (e.g., categories of the instances in the whole image) are used to train an object detector. Many existing methods follow the standard multiple instance learning (MIL) paradigm and have achieved promising performance. However, the lack of deterministic information leads to part domination and missing instances. To address these issues, this paper focuses on identifying and fully exploiting the deterministic information in WSOD. We discover that negative instances (i.e. absolutely wrong instances), ignored in most of the previous studies, normally contain valuable deterministic information. Based on this observation, we here propose a negative deterministic information (NDI) based method for improving WSOD, namely NDI-WSOD. Specifically, our method consists of two stages: NDI collecting and exploiting. In the collecting stage, we design several processes to identify and distill the NDI from negative instances online. In the exploiting stage, we utilize the extracted NDI to construct a novel negative contrastive learning mechanism and a negative guided instance selection strategy for dealing with the issues of part domination and missing instances, respectively. Experimental results on several public benchmarks including VOC 2007, VOC 2012 and MS COCO show that our method achieves satisfactory performance.
\end{abstract}

\section{Introduction}
With the rapid development of computer vision, object detection has attracted widespread attention \cite{fast,faster,yolo}. However, most of these methods rely on strong supervision in instance-level annotation data \cite{VOC,MSCOCO}, which requires tremendous manpower and time consumption. To solve this problem, some weakly supervised object detection (WSOD) methods \cite{WSDDN,OICR,OIM,MIST} have been proposed, where only image-level annotations are required.

Most existing WSOD methods follow the standard Multiple Instance Learning (MIL) pipeline to transform WSOD into a multi-label classification task. However, due to the lack of accurate instance-level annotations, these methods have two constraints. On the one hand, only an image classification task can be constructed to provide supervisory information, resulting in the discriminative part domination issue, whereby it is difficult to determine whether or not a bounding box contains a complete object \cite{WSOD2,MIST}, especially for non-rigid objects such as dogs and cats. On the other hand, existing methods usually sample the top confident proposal to generate pseudo-label for training the instance refinement branch of the system, whilst ignoring other potential objects, leading to missing detections \cite{OIM}.
\begin{figure}[t]
	\centering
	\includegraphics[width=0.47\textwidth]{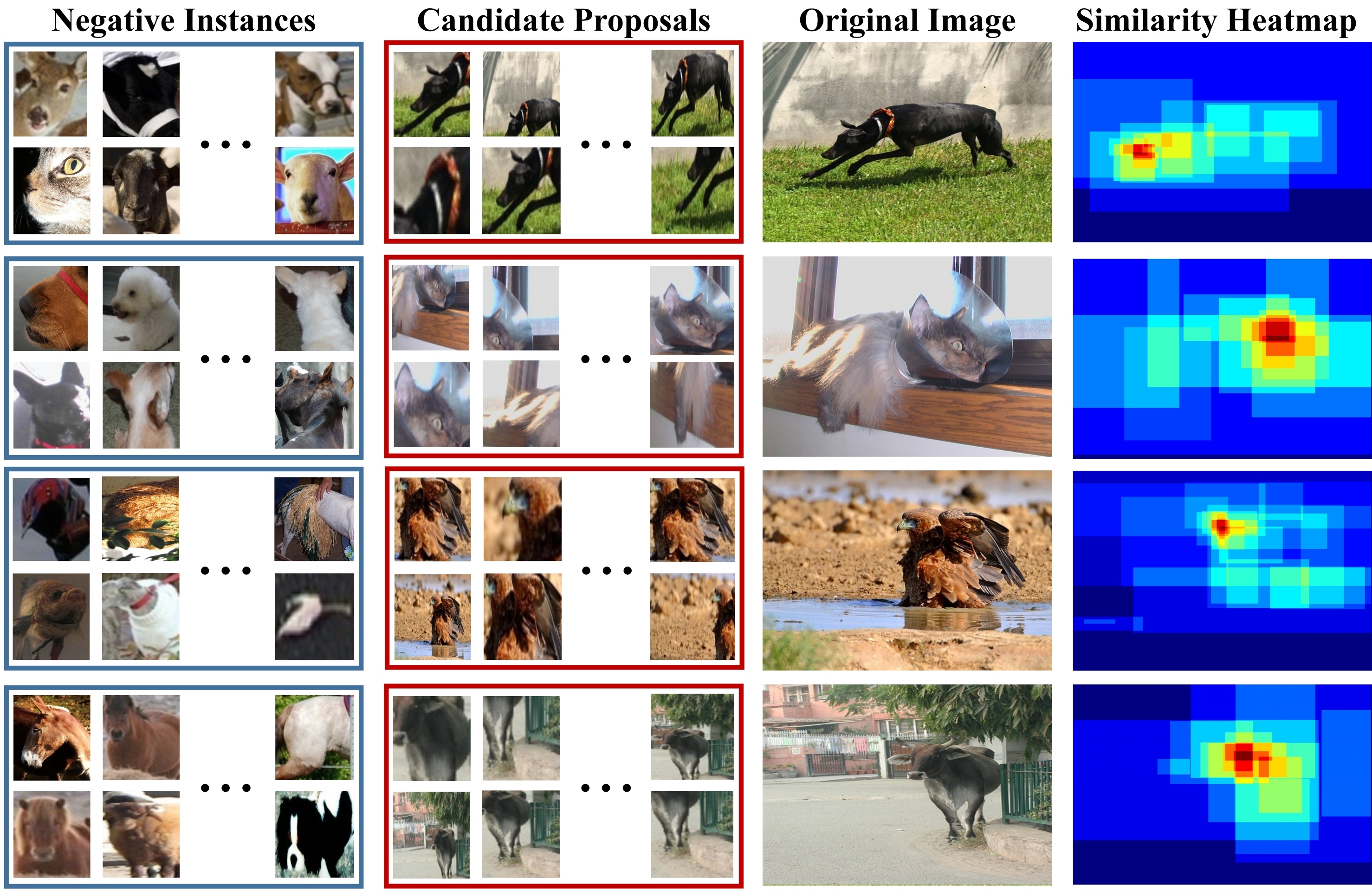} 
	\caption{Visualization of the feature similarities between each proposal and negative instances. The negative instances generated during the training are collected. For each proposal, we calculate the cosine similarity between itself and its category-specific negative instances, then take the maximum as the heatmap score. }
	\label{fig1}
\end{figure}
To address the above issues, Ren et al. \cite{MIST} designed a data-driven spatial dropout block to discover complete objects. Lin et al. \cite{OIM} mined more potential objects for generating pseudo-labels by introducing label propagation on spatial and appearance graphs. Nevertheless, utilizing unreliable supervision tools (i.e. it is uncertain which category each proposal belongs to), the aforementioned limitations still exist.

To handle this challenge, we take a detour from unreliable supervision, aiming to utilize the valuable deterministic information in WSOD to boost system performance. Inspired by the work of \cite{NLNL}, it is straightforward to determine which category the proposal does not belong to. For example, if the current image label is dog, whilst some instances are predicted as cats, it suggests that these instances are absolutely wrong and do not belong to cat, which is a kind of deterministic information. These instances containing deterministic information are referred to as \textbf{Negative Instances}. Figure 1 reveals our interesting discovery that negative instances always contain category-specific discriminative features. As shown in Figure 1, high feature similarities between the object region and its negative instances can be established, while the rest of the object may be ignored. We argue that this ignorance is due to the overfitting of discriminative regions (e.g. the heads of the dog and the bird), the detector tends to produce false predictions in the regions containing category discriminative features. This valuable information of negative instances is referred to as \textbf{Negative Deterministic Information (NDI)}. How to make full use of the NDI may be a key to improve system performance.

To this end, we propose NDI-WSOD to address the above concern taking full advantage of negative instances. Technically speaking, NDI-WSOD consists of a collecting stage and an exploiting stage. In the collecting stage,  we aim to mine the negative instances generated during the training phase and then distill the NDI of them. As mentioned above, these extracted NDI can be treated as the discriminative features of the misclassified category. We here construct a feature bank to store the features for later use.

Afterward, to address the issues of discriminative part domination and missing instances, we utilize the collected NDI and propose a novel negative contrastive learning (NCL) mechanism and a negative guided instance selection (NGIS) strategy in the exploiting stage. The NCL treats the NDI as a template to detect the overfitted discriminative regions and then reduce their impacts, pulling the detector out of the discriminative part domination issue. Different from the activation suppression strategy of the legacy methods that leverage the model’s responses to a single image, the NCL exploits the deterministic information from the entire dataset for better reduction. To solve the missing instances issue, the NGIS selects as many foreground instances as possible for instance refinement, yet avoids unwanted noise by utilizing the collected NDI as a criterion, ensuring both recall and precision.

The contributions of our paper include:
\begin{itemize}
	\item
	We discover negative instances may contain valuable deterministic information in WSOD and propose an effective scheme to collect NDI from these instances online.
	\item Exploiting the collected NDI, we design NCL and NGIS to prevent the WSOD model from the issues of discriminative part domination and missing instances, respectively.
	\item Experiments on VOC 2007, VOC 2012 and MS COCO datasets demonstrate the effectiveness of our proposed NDI-WSOD.
\end{itemize}

\begin{figure*}[t]
	\centering
	\includegraphics[width=0.95\textwidth]{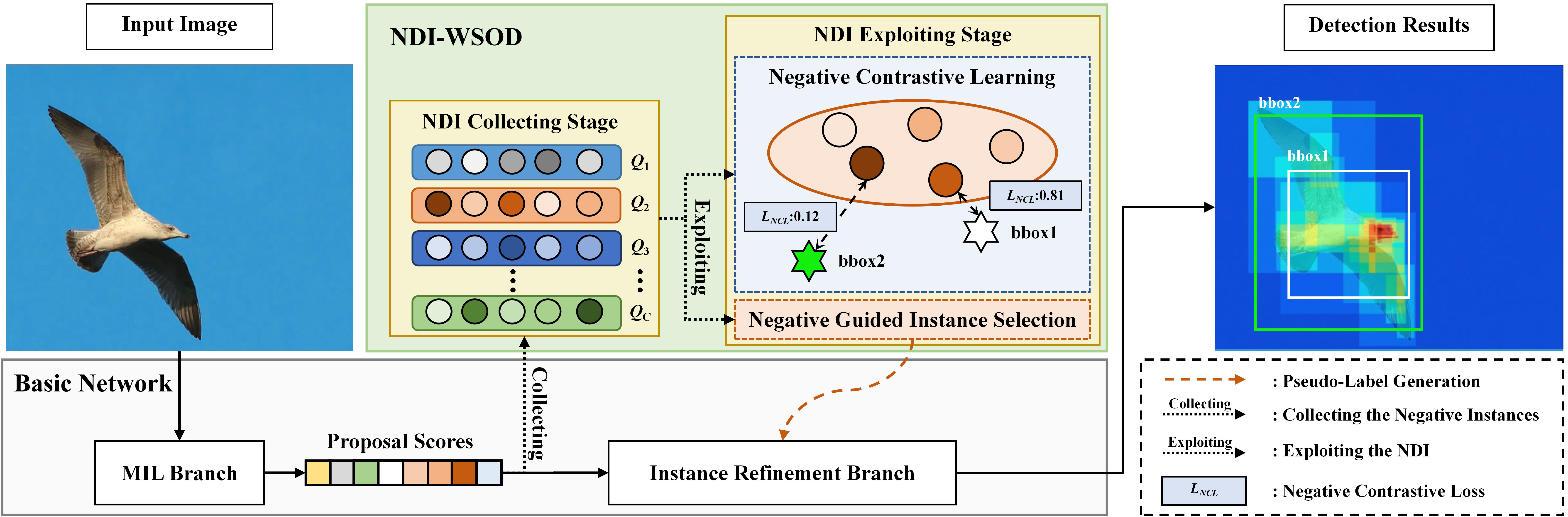} 
	\caption{The architecture of the proposed method. A MIL-branch and an instance refinement branch are used as the basic network. The proposed NDI-WSOD first uses a collecting stage to extract NDI from negative instances generated during the training phase. Based on this, NCL and NGIS are constructed to prevent the model from discriminative part domination and missing instances, respectively.}
	\label{fig2}
\end{figure*}
\section{Related Work}
Existing WSOD methods are usually based on the standard MIL, which treats each image as a bag and a series of pre-computed proposals as instances. WSDDN [Bilen and Vedaldi, 2016] combines convolutional neural networks with MIL, widely used as the basis of future studies. WSDDN-based methods tend to focus on the discriminative regions of a salient object, resulting in incomplete or missing detections. To handle these issues, OICR \cite{OICR} samples the top confident instance to build pseudo-labels that are utilized to train the instance refinement branches for better detection. From the perspective of optimization, C-MIL \cite{C-MIL} smooth the objective function to alleviate the issue of local optimal solutions. Leveraging the feature consistency under different transformations, CASD \cite{CASD} achieves outstanding performance. To enhance the pseudo-label generation, OIM \cite{OIM} aims at mining more positive objects for instance refinement branches by introducing information over spatial and appearance graphs while MIST \cite{MIST} employs a novel multiple instance self-training strategy. Unlike the above methods where only single image information is utilized, IM-CFB \cite{IM-CFB} uses the category-specific features across the dataset as a template to go through instances. Different from the legacy methods, which still rely on unreliable supervision to guide detectors, our proposed NDI-WSOD aims to improve the model by exploiting deterministic information.

Recently, contrastive learning has attracted wide attention in image classification tasks. Moco \cite{MOCO} adopts a dynamic memory module to collect negative samples, which discards the batch size limitation and facilitates unsupervised contrastive learning. SCL \cite{SCL} proposes a supervised contrastive loss in image classification tasks, in which category information is used to construct positive and negative samples. Unlike the above methods that require constructing both positive and negative pairs, BYOL \cite{BYOL} only employs positive samples for constructing a teacher-student model with self-distillation to obtain robust feature representations. Different from the common contrastive learning methods, we here analyse the characteristics of WSOD and then propose a negative contrastive learning mechanism in which only negative samples are adopted to pull the model away from local optima. To the best of our knowledge, this paper is the first in its kind to apply contrastive learning to WSOD.
\section{Proposed Method}
The overall architecture of the proposed method is shown in Figure 2. We employ a MIL branch and an instance refinement branch as the basic network and integrate the proposed NDI-WSOD consisting of a collecting and an exploiting stages into the framework. In the collecting stage, we inspect negative instances generated during the training and extract the NDI from them. In the exploiting stage, based on the collected NDI, a negative contrastive learning (NCL) mechanism and a negative guided instance selection (NGIS) strategy are constructed to deal with discriminative part domination and missing instances, respectively.
\subsection{Preliminaries}
We first build a basic network following the general WSOD framework \cite{WSDDN,OICR}, which consists of a MIL branch and an instance refinement branch. The MIL branch is consistent with WSDDN \cite{WSDDN}, where, given an input image $I$ and its pre-computed proposals $B=\{b_1, b_2, ..., b_N\}$ \cite{SS,MCG}, a corresponding set of proposal features $F=\{f_1, f_2, ..., f_N\}$ are extracted. Then, these features are mapped into proposal scores $S=\{S_{b_1}, S_{b_2}, ..., S_{b_N}\}$. Next, the image classification score on class $c$ is computed through the summation over all the proposals: $\phi_{c}=\sum_{i=1}^{|B|}S^c_{b_i}$. Then, a multi-label classification loss function is imposed as follows:
\begin{equation}
	\mathcal{L}_{{mil}}=-\sum_{c=1}^{C}\left\{y_{c} \log \phi_{c}+\left(1-y_{c}\right) \log \left(1-\phi_{c}\right)\right\}
\end{equation}

For instance refinement branch, we keep the same structure and refinement times as those indicated in \cite{OICR} to obtain precise detections. The branch maps the proposal features to the refined scores through a single-layer classifier. Especially, an extra regression branch is added in the last refinement to regress boxes online \cite{OICR+reg}. The instance refinement branch is trained with pseudo-labels generated by the instance selection strategy of \cite{MIST}, and a multi-task loss is calculated as follows:
\begin{equation}
	\mathcal{L}_{ {ref }}=\mathcal{L}_{ref_{cls}}+\mathcal{L}_{ref_{reg }} .
\end{equation}
where $\mathcal{L}_{ref_{cls}}$ and $\mathcal{L}_{ref_{reg}}$ are the weighted Cross-Entropy and Smooth-L1 loss, respectively, in which the weight is calculated as shown in \cite{OICR}. During the inference phase, we average all the refined scores as the final score for each proposal.
\subsection{NDI-WSOD}
Due to the lack of instance-level supervision, existing WSOD methods often suffer from discriminative part domination and missing instances issues. To handle these two challenges, our NDI-WSOD explores the valuable deterministic information in WSOD, i.e. NDI, and we then propose to exploit the information to boost system performance.
\subsubsection{NDI Collecting Stage}
We first design an NDI collecting stage to collect and extract NDI online. In the NDI collecting stage, our goal is to automatically mine the negative instances and distill the category-specific discriminative features from the negative instances, a kind of deterministic information. Inspired by \cite{IM-CFB}, a feature bank structure is employed to store the extracted NDI. Formally, the collection module contains $C$ queues $\{Q_1, Q_2,..., Q_C\}$, where $C$ denotes the amount of the categories. For the $c$-th category queue, $Q_c = \{(nf^c_1, ns^c_1), (nf^c_2, ns^c_2), ... , (nf^c_L, ns^c_L)\}$, where $L$ is the length of the queue, $nf$ and $ns$ denote the collected features and the corresponding confidence ($ns \in [0, 1]$), respectively. During the training phase, we monitor the proposal scores $S$ from the MIL Branch and identify the negative instances according to the image-level labels.

However, not all the negative instances are valuable. Some negative instances generated by insufficient training of the model only contain noise. To ensure the quality of NDI, we propose the following steps. Firstly, the confidence threshold $\tau$ is set to filter out worthless instances generated due to the flawed model. The retained negative instances can be treated with high confidence as belonging to the misclassified category. These instances usually include a high proportion of discriminative category features. We set an empty queue to collect the proposal features and misclassified-category confidence of these instances. Once full, a confidence-driven momentum update (CMU) strategy is adopted to update the queue for distilling the high-quality NDI in the collected instances.  The CMU first calculates the cosine similarity between the current instance and each feature in its category queue, and then selects the most similar one to update as follows:
\begin{equation}
	nf^c_i \leftarrow \frac{ns^c_i}{ns^c_i+S^{c}_{new}} \cdot nf^c_i  +  \frac{S^{c}_{new}}{ns^c_i+S^{c}_{new}} \cdot f_{new}
\end{equation}
where $f_{new}$ and $S^c_{new}$ are the feature and misclassified-category confidence of the current instance, respectively. The same operation is performed for $ns^c_i$. Following this style, the high-quality NDI is further extracted.
\begin{figure}[t]
	\centering
	\includegraphics[width=0.3\textwidth]{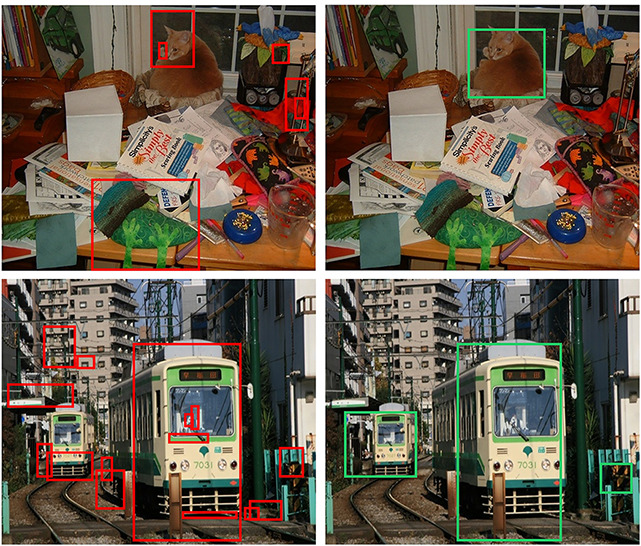} 
	\caption{Visualization of the instance selection strategy. The image-level labels of the 1st and 2nd rows are cat and train, respectively. The baseline (the left image of the pair) is inferior to our NGIS in the avoidance of noise.}
	\label{fig3}
\end{figure}

In addition, we design an auxiliary loss, namely negative instance cross entropy (NICE) loss, to eliminate the noisy instances that are of insufficient training. With only image-level annotations, it is difficult to determine whether or not an instance belongs to the current category. However, it is easy to understand why an instance does not belong to certain categories. Hence, we utilize this deterministic information to construct a NICE loss for the misclassified instances as follows:
\begin{equation}
	\mathcal{L}_{nice}=- \sum_{j=1}^{|B|} \mathbb{I}(S^{neg}_{b_j}>\tau) \cdot y^{neg}_{b_j} \log (1-S^{neg}_{b_j})
\end{equation}
where $|B|$ denotes the number of the instances, $y^{neg}_{b_j}$ denotes the misclassified category of the $j$-th instance and $S^{neg}_{b_j}$ refers to as the corresponding confidence. The proposed NICE loss enables the misclassified instances to be optimized, reducing noisy instances caused by insufficient training. With the help of the above steps, the high-quality NDI can be obtained.
\subsubsection{NDI Exploiting Stage}
Leveraging the collected NDI, we here propose a negative contrastive learning mechanism and a negative guided instance selection strategy to alleviate discriminative part domination and missing instances issues, respectively.

\textbf{(1) Negative Contrastive Learning}: Since only a multi-label classification task can be constructed to provide supervisory information, only the regions that are beneficial to the classification task are attended, which is called the discriminative part domination issue. It is intuitive for someone to force the model to punish the regions that are overly focused. Accurate identification of the overfitted regions and restriction of their impacts are critical.

We observe that the NDI that contain overfitted region features can be treated as templates to match the overfitted regions. To do so, we follow a contrastive learning paradigm, which utilizes each instance and its corresponding category NDI to construct dissimilar sample pairs while driving them distant in the representation space, thus guiding the model to escape from part domination. Given a series of instances, we first filter out worthless instances with the threshold $\tau$ like the first step in the collecting stage. Next, we calculate the distance between each instance and its corresponding NDI. For simplicity, cosine similarity is adopted for the computation. Then, the NCL loss can be represented as follows:
\begin{equation}
	\mathcal{L}_{ncl}=\alpha \cdot \sum_{j=1}^{|B|} {{ns}_{i*}^{c_{b_j}}} \cdot \max _{i}\Big( \frac{f_{b_j} \cdot {nf}_{i}^{c_{b_j}}}{||f_{b_j}|| \cdot ||{nf}_{i}^{c_{b_j}}||}\Big)
\end{equation}
where $\alpha$ is a trade-off factor, $c_{b_j}$ and $f_{b_j}$ are the category and proposal features of the $j$-th proposal respectively, and $i*$ is the key of the closest NDI. ${ns}_{i*}^{c_{b_j}}$ reveals the proportion of the discriminative features in the NDI, and the feature similarity between the current instance and the NDI represents its probability of belonging to discriminative regions. High similarity means that the instance should be significantly punished. As illustrated in Figure 2, bbox1 contains a higher proportion of the discriminative region than bbox2, thus a large loss is imposed forcing the model to look for the more complete object bbox2.

With the help of NCL, more complete objects can be detected. In contrast to the legacy methods that punish the regions by leveraging the model’s response to a single image, we here exploit the NDI of the entire dataset to accurately suppress discriminative regions, whereby solving the discriminative part domination issue.

\textbf{(2) Negative Guided Instance Selection}: The instance refinement branch is widely used in the existing methods for improving system outcomes. Training this branch relies on pseudo-label generation, where positive instance selection plays a key role. Existing methods usually select the top confidence object as a positive instance, whilst ignoring other potential objects, resulting in missing instances \cite{OICR}. Or, one selects the top 15\% of the proposals as positive instances, which can secure the recall but introduces more noise \cite{MIST}. To this end, based on the available NDI, we design an NGIS strategy to deal with the problem.

Following the process of elimination, NGIS has been designed to select a large set of candidates and then filter out the noisy ones. For the former, NGIS uses the way similar to \cite{MIST} that selects the top 15\% of the non-overlapping instances as candidates. Since these candidates usually include much background noise, we then utilize the NDI to remove the noisy ones. Considering the prior knowledge that positive instances usually contain the features of discriminative regions, we employ the NDI as a template to scrutinize each candidate. Specifically, we calculate the max cosine distance $dis^{b_i}_{cos}$ between each instance and the NDI. Then, we filter out the noisy instances with the criteria: $dis^{b_i}_{cos}>\beta \cdot \frac{1}{|B|} \cdot \sum_{j=1}^{|B|} dis^{b_j}_{cos}$. As illustrated in Figure 3, in comparison with \cite{MIST}, NGIS can filter out the noise in complex environments and maintain a large number of positive instances, thus addressing missing instances.
\begin{figure*}[t]
	\centering
	\includegraphics[width=0.9\textwidth]{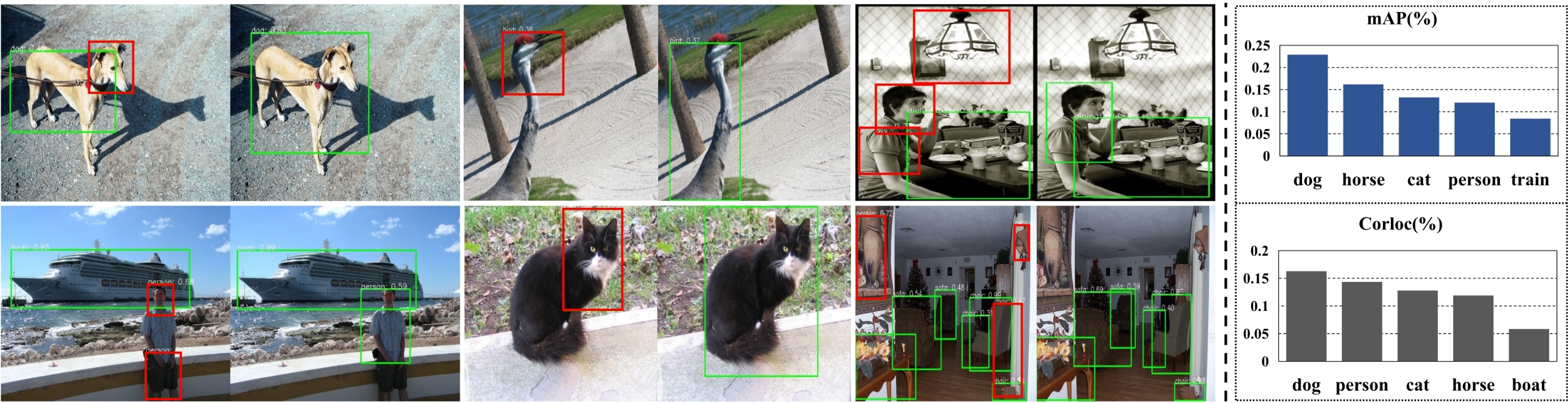} 
	\caption{Qualitative results of our NDI-WSOD (the right image in the pair) and the baseline (the left image in the pair) sit on the left hand side of the dashed line, where the failing cases are highlighted with red boxes. Top-5 categories with the largest performance improvements are exhibited on the right hand side.}
	\label{fig4}
\end{figure*}
\begin{table}[t]
	\centering
	\small
	\begin{tabular}{lcc}
		\toprule
		Method & mAP(\%)& CorLoc(\%) \\
		\midrule
		OICR \cite{OICR}& 41.2& 60.6 \\
		PCL \cite{PCL}& 43.5& 62.7 \\
		OIM \cite{OIM}& 50.1& 67.2 \\
		C-MIL \cite{C-MIL}& 50.5& 65.0 \\
		WSOD$^2$ \cite{WSOD2}& 53.6& 69.5 \\
		IM-CFB \cite{IM-CFB}& 54.3& 70.7 \\
		MIST \cite{MIST}& 54.9& 68.8 \\
		\midrule
		NDI-WSOD & \textbf{56.8}& \textbf{71.0} \\
		\bottomrule
	\end{tabular}
	\caption{Comparison with the other state-of-the-arts on the VOC 2007 dataset.}
	\label{table1}
\end{table}

\section{Experiments and Analysis}
\subsection{Datasets and Evaluation Metrics}
Following the other state-of-the-art methods, we evaluate our NDI-WSOD on several challenging object detection benchmarks: VOC 2007, VOC 2012 \cite{VOC} and MS COCO \cite{MSCOCO}. For the VOC datasets, we employ the trainval set for training and evaluate the system performance on the test set. Mean average precision (mAP) and Correct Localization (CorLoc) are adopted to evaluate the overall performance and the localization power of the model, respectively. For MS COCO, we employ the train2014 set for training and evaluate the system performance on the val2014 set, where mAP and mAP$_{[.5:.05:.95]}$ are employed to evaluate the overall performance. In addition, mAP$_s$, mAP$_m$, and mAP$_l$ are used to evaluate the system performance for small, medium, and large objects respectively.
\subsection{Implementation Details}
For a fair comparison, Selective Search \cite{SS} and MCG \cite{MCG} are used for VOC \cite{VOC} and MS COCO \cite{MSCOCO} to generate proposals, respectively. Then, the whole model is trained on a single NVIDIA GeForce GTX 2080Ti with 11-GB GPU memory, and the batch size is set to 4. SGD with an initial learning rate of 0.0005, weight decay of 0.0005 and momentum of 0.9 are used to optimize the model. The overall iteration numbers are set to 35,000, 70,000, 210,000 for VOC 2007, VOC 2012, and MS COCO, respectively. The corresponding learning rate is decreased by a factor of 10 at the 30,000th, 60,000th,  100,000th step, respectively. For data augmentation, we keep the same strategy as the state of the art methods, including multi-scale training and random horizontal flipping. The length $L$ and threshold $\tau$ in the collecting stage are set to 5 and 0.05, respectively. In the exploiting stage, the factor $\alpha$ and $\beta$ are set to 0.3 and 3, respectively. During the training phase, $\mathcal{L}_{ {total }}=\mathcal{L}_{ {mil }}+\mathcal{L}_{ {ref }}+\mathcal{L}_{ {nice }}+\mathcal{L}_{ {ncl }}$.

\begin{table}[t]
	\centering
	\small
	\begin{tabular}{lcc}
		\toprule
		Method & mAP(\%)&CorLoc(\%) \\
		\midrule
		OICR \cite{OICR}& 37.9 & 62.1 \\
		PCL \cite{PCL}& 40.6 & 63.2 \\
		OIM \cite{OIM}& 45.3 & 67.1 \\
		C-MIL \cite{C-MIL}& 46.7 & 67.4 \\
		WSOD$^2$ \cite{WSOD2}& 47.2 & 71.9 \\
		IM-CFB \cite{IM-CFB}& 49.4 & 69.6 \\
		MIST \cite{MIST}& 52.1 & 70.9 \\
		\midrule
		NDI-WSOD& \textbf{53.9} & \textbf{72.2} \\
		\bottomrule
	\end{tabular}
	\caption{Comparison with the other state-of-the-arts on the VOC 2012 dataset.} 
	\label{table2}
\end{table}
\begin{table}[t]
	\centering
	\small
	\resizebox{0.98\columnwidth}{!}{
		\begin{tabular}{l|ccccc}
			\toprule
			Method&mAP$_{[.5:.05:.95]}$&mAP&mAP$_s$&mAP$_m$&mAP$_l$ \\
			\midrule
			PCL&8.5& 19.4&-&-&- \\
			WSOD$^2$&10.8& 22.7&-&-&- \\
			MIST&11.4& 24.3&3.6&12.2&17.6 \\
			\midrule
			NDI-WSOD&\textbf{12.1}&\textbf{26.2}& \textbf{3.7}&\textbf{13.2}&\textbf{19.3}\\
			\bottomrule
		\end{tabular}
	}
	\caption{Comparison with the other state-of-the-arts on the MS COCO dataset.}
	\label{table3}
\end{table}

\subsection{Comparison with State-of-the-arts}
We evaluate our method on VOC 2007, VOC 2012 and MS COCO, in comparison with several state-of-the-art methods. Table 1 shows the overall performance on the VOC 2007 dataset. The results show that NDI-WSOD achieves 56.8\% mAP and 71.0\% CorLoc, which significantly outperforms the other methods. Especially, in comparison with the state-of-the-art method \cite{MIST}, NDI-WSOD achieves a significant improvement with 1.9\% mAP and 2.2\% CorLoc, respectively, demonstrating the superiority of our method.

To demonstrate the robustness of NDI-WSOD, we also evaluate our method on VOC 2012. Table 2 shows the detection and localization results on the VOC 2012 test and trainval sets, respectively. As can be seen, our NDI-WSOD achieves 53.9\% mAP\footnote{\url{http://host.robots.ox.ac.uk:8080/anonymous/OGCPWE.html}} and 72.2\% CorLoc, which outperforms all the other methods with at least 1.8\% increments in terms of mAP.

To verify that NDI-WSOD has improved the detection IoU, we report the detection results on the large dataset MS COCO in Table 3. It is witnessed that NDI-WSOD outperforms the other methods with at least 0.7\%/1.9\% increment in terms of mAP$_{[.5:.05:.95]}$ and mAP, respectively. In addition, we evaluate the system performance with different object sizes. In comparison to \cite{MIST}, NDI-WSOD performs better over different object sizes, especially for large objects that are likely to be partially detected, showing that our method achieves the expectation.

\subsection{Ablation Study}

\begin{table}[t]
	\centering
	\small
	\begin{tabular}{l|c|c}
		\toprule
		Method & mAP(\%)&CorLoc(\%) \\
		\midrule
		OICR$\dagger$& 41.8 &  58.7\\
		OICR$\dagger$+NCL& 44.6 (+2.8) & 61.8 (+3.1) \\
		PCL$\dagger$& 46.2&  64.4\\
		PCL$\dagger$+NCL& 48.8 (+2.6) & 66.8 (+2.4) \\
		\midrule
		Baseline& 52.9 & 68.5\\
		Baseline+NCL& 56.0 (+3.1) & 70.4 (+1.9) \\
		Baseline+NGIS& 54.7 (+1.8) & 69.6 (+1.1) \\
		NDI-WSOD& \textbf{56.8 (+3.9)} & \textbf{71.0 (+2.5)} \\
		\bottomrule
	\end{tabular}
	\caption{Ablation study on the VOC 2007 dataset ($\dagger$ indicates our implementation).}
	\label{table4}
\end{table}

\subsubsection{Effect of each component}
We perform an ablation study on VOC 2007 to verify the effectiveness of each component in NDI-WSOD. We first introduce the pseudo label generation strategy of \cite{MIST} into our basic network as the baseline. As shown in Table 4, in comparison with the baseline, adding the NCL yields a salient improvement, which brings 3.1\%/1.9\% improvements on mAP/CorLoc. On the other hand, employing the NGIS brings 1.8\%/1.1\% improvements on mAP/CorLoc, which indicates sampling more potential objects is helpful for performance improvement. Integrating all the modules, the system achieves the best performance, significantly better than the baseline with 3.9\%/2.5\% improvements on mAP/CorLoc.

To verify the generalization ability of NDI-WSOD, we insert the NCL mechanism into the classical OICR \cite{OICR} and PCL \cite{PCL}. The results respectively observe 2.8\%/3.1\% and 2.6\%/2.4\% improvements on mAP/CorLoc. Moreover, we visualize the detection results of NDI-WSOD in Figure 4, which shows NDI-WSOD achieves a significant improvement on non-rigid objects (e.g. cats, dogs, and persons) and also work satisfactorily on complex scenes, further confirming the effectiveness of our proposed method.
\begin{table}[t]
	\centering
	\small
	\begin{tabular}{l|c|c}
		\toprule
		Update strategy &Length &mAP(\%) \\
		\midrule
		FIFO& 5& 53.9 \\
		CMU& 5 & 54.9 \\
		CMU+$\mathcal{L}_{ {nice }}$& 5 &  \textbf{56.8}\\
		\midrule
		CMU+$\mathcal{L}_{ {nice }}$& 1&  54.3\\
		CMU+$\mathcal{L}_{ {nice }}$& 3 & 54.6\\
		CMU+$\mathcal{L}_{ {nice }}$& 7 & 55.8\\
		CMU+$\mathcal{L}_{ {nice }}$& 9& 55.1\\
		\bottomrule
	\end{tabular}
	\caption{Constructing strategy of NDI collecting stage on the VOC 2007 test set in terms of mAP (\%).}
	\label{table5}
\end{table}

\subsubsection{Construction of the NDI collecting stage}
A simple yet effective First-In-First-Out (FIFO) strategy is selected as the baseline in this experiment. As shown in Table 5, applying the CMU strategy brings 1.0\% improvement on mAP. Adding $\mathcal{L}_{ {nice }}$ improves the system performance with 1.9\% mAP, which indicates that suppressing the noisy instances can enhance the quality of NDI. Then, we analyze the effect of $L$. As we can see, in contrast to a larger $L$, a smaller $L$ will reduce the noise tolerance, resulting in a significant system performance decrease.
\begin{figure}[t]
	\centering
	\includegraphics[width=0.45\textwidth]{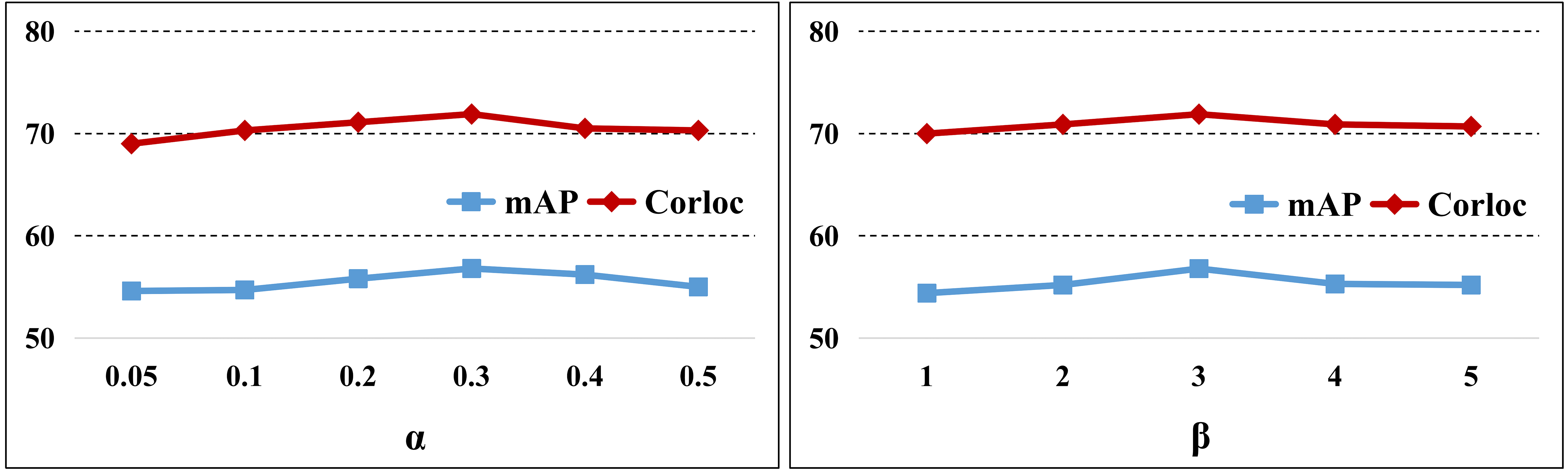} 
	\caption{Detection results with different impact factors on the VOC 2007 dataset.}
	\label{fig5}
\end{figure}
\subsubsection{Impact factor}
For loss factor $\alpha$ in NCL, Figure 5 shows the experimental results, where $\alpha$=0.3 leads to the best outcome, as it makes the best balance between the losses. For factor $\beta$ in NGIS, the results indicate $\beta$=3 is the best trade-off. A larger $\beta$ normally results in object missing. On the contrary, a smaller $\beta$ is not sufficient to filter the noise. It is worth noting that our NDI-WSOD is less sensitive to the variations of the newly introduced hyperparameters, and thus can be effectively applied in the real world practice.
\section{Conclusion}
In this paper, we proposed a novel negative deterministic information (NDI) based method, namely NDI-WSOD. We have discovered that negative instances usually contain valuable NDI and then designed a collecting stage to distill the negative instances. Utilizing this NDI, we proposed an exploiting stage consisting of a negative contrastive learning (NCL) mechanism and a negative guided instance selection (NGIS) strategy to deal with part domination and missing instances, respectively. Experimental results conducted on VOC 2007, VOC 2012 and MS COCO demonstrated the effectiveness of our proposed NDI-WSOD.

\newpage
\bibliographystyle{named}
\bibliography{NDI}

\begin{thebibliography}{}

\bibitem[\protect\citeauthoryear{Arbel{\'a}ez \bgroup \em et al.\egroup
  }{2014}]{MCG}
Pablo Arbel{\'a}ez, Jordi Pont-Tuset, Jonathan~T Barron, Ferran Marques, and
  Jitendra Malik.
\newblock {Multiscale combinatorial grouping}.
\newblock In {\em Proceedings of the IEEE Conference on Computer Vision and
  Pattern Recognition {(CVPR)}}, pages 328--335, 2014.

\bibitem[\protect\citeauthoryear{Bilen and Vedaldi}{2016}]{WSDDN}
Hakan Bilen and Andrea Vedaldi.
\newblock {Weakly supervised deep detection networks}.
\newblock In {\em Proceedings of the IEEE conference on computer vision and
  pattern recognition {(CVPR)}}, pages 2846--2854, 2016.

\bibitem[\protect\citeauthoryear{Everingham \bgroup \em et al.\egroup
  }{2010}]{VOC}
Mark Everingham, Luc Van~Gool, Christopher~KI Williams, John Winn, and Andrew
  Zisserman.
\newblock The pascal visual object classes (voc) challenge.
\newblock {\em International journal of computer vision {(IJCV)}},
  88(2):303--338, 2010.

\bibitem[\protect\citeauthoryear{Girshick}{2015}]{fast}
Ross. Girshick.
\newblock {Fast r-cnn}.
\newblock In {\em Proceedings of the IEEE international conference on computer
  vision {(ICCV)}}, pages 1440--1448, 2015.

\bibitem[\protect\citeauthoryear{Grill \bgroup \em et al.\egroup }{2020}]{BYOL}
Jean-Bastien Grill, Florian Strub, Florent Altch{\'e}, Corentin Tallec,
  Pierre~H Richemond, Elena Buchatskaya, Carl Doersch, Bernardo~Avila Pires,
  Zhaohan~Daniel Guo, Mohammad~Gheshlaghi Azar, et~al.
\newblock {Bootstrap your own latent: A new approach to self-supervised
  learning}.
\newblock In {\em Advances in neural information processing systems
  {(NeurIPS)}}, 2020.

\bibitem[\protect\citeauthoryear{He \bgroup \em et al.\egroup }{2020}]{MOCO}
Kaiming He, Haoqi Fan, Yuxin Wu, Saining Xie, and Ross Girshick.
\newblock {Momentum contrast for unsupervised visual representation learning}.
\newblock In {\em Proceedings of the IEEE Conference on Computer Vision and
  Pattern Recognition {(CVPR)}}, pages 9729--9738, 2020.

\bibitem[\protect\citeauthoryear{Huang \bgroup \em et al.\egroup }{2020}]{CASD}
Zeyi Huang, Yang Zou, Vijayakumar Bhagavatula, and Dong Huang.
\newblock {Comprehensive attention self-distillation for weakly-supervised
  object detection}.
\newblock In {\em Advances in neural information processing systems
  {(NeurIPS)}}, 2020.

\bibitem[\protect\citeauthoryear{Khosla \bgroup \em et al.\egroup }{2020}]{SCL}
Prannay Khosla, Piotr Teterwak, Chen Wang, Aaron Sarna, Yonglong Tian, Phillip
  Isola, Aaron Maschinot, Ce~Liu, and Dilip Krishnan.
\newblock {Supervised contrastive learning}.
\newblock In {\em Advances in neural information processing systems
  {(NeurIPS)}}, 2020.

\bibitem[\protect\citeauthoryear{Kim \bgroup \em et al.\egroup }{2019}]{NLNL}
Youngdong Kim, Junho Yim, Juseung Yun, and Junmo Kim.
\newblock {Nlnl: Negative learning for noisy labels}.
\newblock In {\em Proceedings of the IEEE International Conference on Computer
  Vision {(ICCV)}}, pages 101--110, 2019.

\bibitem[\protect\citeauthoryear{Lin \bgroup \em et al.\egroup }{2014}]{MSCOCO}
Tsung-Yi Lin, Michael Maire, Serge Belongie, James Hays, Pietro Perona, Deva
  Ramanan, Piotr Doll{\'a}r, and C~Lawrence Zitnick.
\newblock {Microsoft coco: Common objects in context}.
\newblock In {\em European conference on computer vision {(ECCV)}}, pages
  740--755, 2014.

\bibitem[\protect\citeauthoryear{Lin \bgroup \em et al.\egroup }{2020}]{OIM}
Chenhao Lin, Siwen Wang, Dongqi Xu, Yu~Lu, and Wayne Zhang.
\newblock {Object instance mining for weakly supervised object detection}.
\newblock In {\em Proceedings of the AAAI Conference on Artificial Intelligence
  {(AAAI)}}, pages 11482--11489, 2020.

\bibitem[\protect\citeauthoryear{Redmon \bgroup \em et al.\egroup
  }{2016}]{yolo}
Joseph Redmon, Santosh Divvala, Ross Girshick, and Ali Farhadi.
\newblock {You only look once: Unified, real-time object detection}.
\newblock In {\em Proceedings of the IEEE conference on computer vision and
  pattern recognition {(CVPR)}}, pages 779--788, 2016.

\bibitem[\protect\citeauthoryear{Ren \bgroup \em et al.\egroup }{2015}]{faster}
Shaoqing Ren, Kaiming He, Ross Girshick, and Jian Sun.
\newblock {Faster r-cnn: Towards real-time object detection with region
  proposal networks}.
\newblock In {\em Advances in neural information processing systems
  {(NeurIPS)}}, pages 91--99, 2015.

\bibitem[\protect\citeauthoryear{Ren \bgroup \em et al.\egroup }{2020}]{MIST}
Zhongzheng Ren, Zhiding Yu, Xiaodong Yang, Ming-Yu Liu, Yong~Jae Lee,
  Alexander~G Schwing, and Jan Kautz.
\newblock {Instance-aware, context-focused, and memory-efficient weakly
  supervised object detection}.
\newblock In {\em Proceedings of the IEEE Conference on Computer Vision and
  Pattern Recognition {(CVPR)}}, pages 10598--10607, 2020.

\bibitem[\protect\citeauthoryear{Tang \bgroup \em et al.\egroup }{2017}]{OICR}
Peng Tang, Xinggang Wang, Xiang Bai, and Wenyu Liu.
\newblock {Multiple instance detection network with online instance classifier
  refinement}.
\newblock In {\em Proceedings of the IEEE conference on computer vision and
  pattern recognition {(CVPR)}}, pages 2843--2851, 2017.

\bibitem[\protect\citeauthoryear{Tang \bgroup \em et al.\egroup }{2018}]{PCL}
Peng Tang, Xinggang Wang, Song Bai, Wei Shen, Xiang Bai, Wenyu Liu, and Alan
  Yuille.
\newblock Pcl: Proposal cluster learning for weakly supervised object
  detection.
\newblock {\em IEEE transactions on pattern analysis and machine intelligence
  {(TPAMI)}}, 42(1):176--191, 2018.

\bibitem[\protect\citeauthoryear{Uijlings \bgroup \em et al.\egroup
  }{2013}]{SS}
Jasper~RR Uijlings, Koen~EA Van De~Sande, Theo Gevers, and Arnold~WM Smeulders.
\newblock Selective search for object recognition.
\newblock {\em International journal of computer vision {(IJCV)}},
  104(2):154--171, 2013.

\bibitem[\protect\citeauthoryear{Wan \bgroup \em et al.\egroup }{2019}]{C-MIL}
Fang Wan, Chang Liu, Wei Ke, Xiangyang Ji, Jianbin Jiao, and Qixiang Ye.
\newblock {C-mil: Continuation multiple instance learning for weakly supervised
  object detection}.
\newblock In {\em Proceedings of the IEEE Conference on Computer Vision and
  Pattern Recognition {(CVPR)}}, pages 2199--2208, 2019.

\bibitem[\protect\citeauthoryear{Yang \bgroup \em et al.\egroup
  }{2019}]{OICR+reg}
Ke~Yang, Dongsheng Li, and Yong Dou.
\newblock {Towards precise end-to-end weakly supervised object detection
  network}.
\newblock In {\em Proceedings of the IEEE Conference on Computer Vision and
  Pattern Recognition {(CVPR)}}, pages 8372--8381, 2019.

\bibitem[\protect\citeauthoryear{Yin \bgroup \em et al.\egroup }{2021}]{IM-CFB}
Yufei Yin, Jiajun Deng, Wengang Zhou, and Houqiang Li.
\newblock {Instance Mining with Class Feature Banks for Weakly Supervised
  Object Detection}.
\newblock In {\em Proceedings of the AAAI Conference on Artificial Intelligence
  {(AAAI)}}, pages 3190--3198, 2021.

\bibitem[\protect\citeauthoryear{Zeng \bgroup \em et al.\egroup }{2019}]{WSOD2}
Zhaoyang Zeng, Bei Liu, Jianlong Fu, Hongyang Chao, and Lei Zhang.
\newblock {Wsod2: Learning bottom-up and top-down objectness distillation for
  weakly-supervised object detection}.
\newblock In {\em Proceedings of the IEEE international conference on computer
  vision {(ICCV)}}, pages 8292--8300, 2019.

\end{thebibliography}

\end{document}